\pdfoutput=1
\documentclass{article}
\usepackage{iclr2027_conference,times}
\iclrfinalcopy
\usepackage[T1]{fontenc}
\usepackage[utf8]{inputenc}

\setcitestyle{authoryear,round,citesep={;},aysep={,},yysep={;}}

\usepackage{amsmath,amsfonts,bm}

\usepackage{hyperref}
\usepackage{url}
\usepackage{booktabs}
\usepackage{graphicx}
\AddToHook{env/tabular/before}{\par\noindent}
\usepackage{placeins}
\usepackage{xcolor}

\title{ALIVE: Interaction-Aligned Object Insertion for First-Frame-Guided Video Editing}

\author{%
  Zhenghong Zhou$^{1}$\thanks{The implementation and experiments reported in this paper were carried out by Zhenghong Zhou at the University of Rochester.} \quad Zhe Lin$^{2}$ \quad
  Jiebo Luo$^{1,*}$ \quad Yuqian Zhou$^{2,*}$ \\
  $^{1}$University of Rochester \qquad $^{2}$Adobe Research
  \quad {\small $^{*}$Advising authors.} \\
  {\small \href{https://real-time-video-research.github.io/alive/}{Project page}}
}
\hypersetup{%
  hidelinks,
  pdftitle={ALIVE: Interaction-Aligned Object Insertion for First-Frame-Guided Video Editing},
  pdfauthor={Zhenghong Zhou, Zhe Lin, Jiebo Luo, Yuqian Zhou},
  pdfsubject={First-frame-guided video object insertion},
  pdfkeywords={video editing, object insertion, interaction, vision-language model}
}

\begin{document}

\setcounter{footnote}{1}
\maketitle
\fancyhead{}
\renewcommand{\headrulewidth}{0pt}
\vspace{-5mm}
\begin{figure}[!h]
    \centering
    \includegraphics[width=\linewidth,height=0.39\textheight,keepaspectratio]{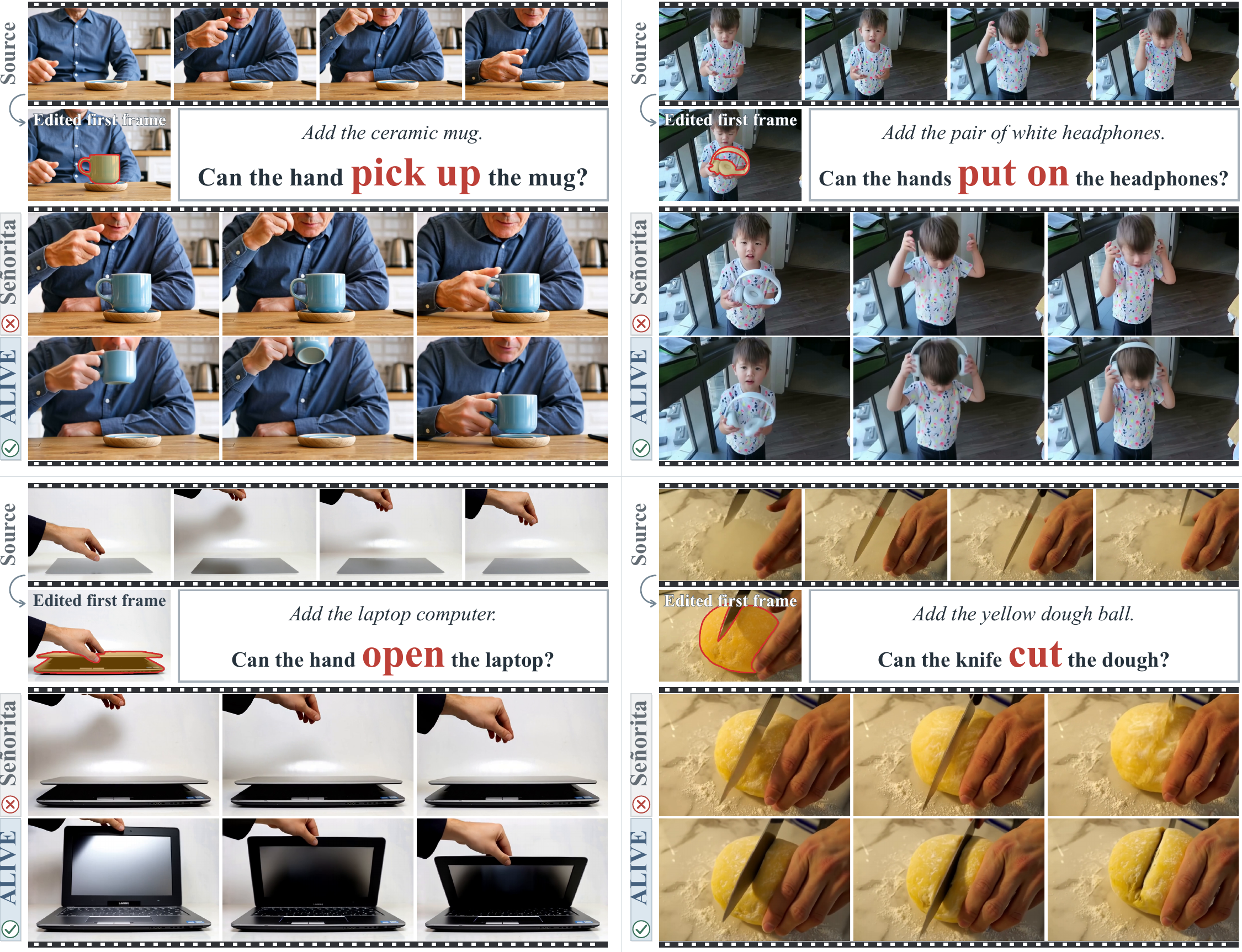}
    \caption{\textbf{Bringing inserted objects to life.} Each panel shows the source video, edited first frame and object-only instruction, followed by corresponding results from Se\~norita and ALIVE. ALIVE enables added objects to be picked up, worn, opened, or cut in response to source actions.}
    \label{fig:teaser}
    \vspace{-1.5mm}
\end{figure}

\begin{abstract}
\small
\vspace{-1.2mm}
Current video editors can insert objects but often struggle to make them participate in interactions such as being picked up or manipulated.
We introduce ALIVE, a framework that makes inserted objects ``alive'' through coherent interactions with the source video's contents, using an edited first frame and an instruction naming only the added object.
We curate 35,800 editing pairs combining 3D-rendered, model-generated, and real-world videos with general editing pairs from ROSE.
Each pair differs in the target object's presence while preserving the surrounding action, teaching editors coordinated object behavior and source preservation.
We further train a vision-language model (VLM) to predict interaction guidance from the same inputs.
We introduce the ALIVE-interaction benchmark to assess interaction fidelity, source preservation, and visual coherence using a unified VLM-based protocol, and evaluate on the general video object insertion benchmark.
Without VLM guidance, ALIVE improves Overall over the strongest evaluated baseline by 43.9\% and 4.4\% on the two benchmarks, respectively.
VLM-predicted guidance further improves the ALIVE-interaction score by 0.95 points without additional user inputs.
\end{abstract}

\section{Introduction}
\label{sec:introduction}
\vspace{-1.5mm}

Recent advances in video generation and editing have enabled high-quality results across diverse tasks, including object insertion with realistic shadows and reflections~\citep{liu2025generative,fu2026effecterase}. However, making these objects respond coherently to source actions remains challenging. In Figure~\ref{fig:teaser}, an added mug should move with the hand that lifts it, and added dough should separate as a knife cuts through it. Our goal is to make inserted objects ``alive'': not merely visible in the video, but part of its world, responding to the actions unfolding around them.

This capability poses two challenges: data construction and interaction modeling. Interaction editing pairs are scarce: their construction requires finding interaction videos and changing the target object's presence while preserving the surrounding scene and action. For modeling, the source video provides action cues but does not directly show the added object's response. The editor must infer the object's evolving motion or state as these actions unfold.

\textbf{ALIVE} addresses these challenges, combining interaction-focused editing data with a vision-language model (VLM) that infers how added objects should respond to source actions. We curate 35,800 editing pairs combining 3D-rendered, model-generated, and real-world videos with general editing pairs from ROSE. These sources cover varied interactions and visual domains, supporting general object insertion. Quality checks exclude implausible interactions and unintended changes to the surrounding scene. We annotate multi-level interaction descriptions, including temporally localized prompts, to study which guidance benefits editing and supervise VLM prediction.

Using these data, we train a first-frame-guided diffusion editor~\citep{ouyang2024i2vedit,liu2025generative}. Users provide a source video, an edited first frame, and an object-only instruction.
The edited frame fixes the object's initial appearance and placement, leaving subsequent behavior to follow source actions.
We train the VLM to predict temporally localized interaction prompts from the same inputs, without user-provided interaction descriptions, per-frame masks, or trajectories.

We introduce the ALIVE-interaction benchmark to test whether added objects participate coherently in source actions while preserving the surrounding content. The general video object insertion benchmark further assesses broader editing quality. Without VLM guidance, ALIVE improves Overall scores over the strongest evaluated baseline by 43.9\% and 4.4\% on the two benchmarks, respectively. Predicted temporal prompt guidance further improves interaction performance. Ablations examine contributions from data composition and interaction guidance.

Our contributions are:
\begin{itemize}
    \vspace{-0.5mm}
    \item A dataset of 35,800 video editing pairs combining interaction-focused and general insertion data, with complementary visual sources and multi-level prompt annotations.
    \vspace{-0.5mm}
    \item A diffusion editing framework that uses a VLM to predict interaction descriptions from the original inputs and aligns their guidance with the corresponding video intervals.
    \vspace{-0.5mm}
    \item We introduce the ALIVE-interaction benchmark and show that ALIVE outperforms baselines on interaction and general insertion, supported by ablations on data and guidance.
\end{itemize}
\vspace{-2mm}

\section{Related Work}
\label{gen_inst}
\label{sec:related_work}
\vspace{-1mm}
\noindent\textbf{Video editing and paired video.} EditVerse, UNIC, and VACE unify diverse video editing tasks through shared conditioning interfaces~\citep{ju2026editverse,ye2025unic,jiang2025vace}. Goku-Edit and OpenVE-Edit support instruction-guided editing~\citep{liang2026goku,he2025openve}, while Kiwi-Edit unifies instruction and reference guidance~\citep{lin2026kiwi}. First-frame-guided methods propagate initial visual edits through source videos. AnyV2V and I2VEdit build on pretrained image-to-video models~\citep{ku2024anyv2v,ouyang2024i2vedit}, while LoRA-Edit uses mask-aware adaptation~\citep{gao2026controllable}. Se\~norita, GenProp, and PropFly learn editing or propagation from paired or synthetic supervision~\citep{zi2025senorita,liu2025generative,seo2026propfly}. PISCO and NovaEdit support sparse-keyframe-guided video editing~\citep{gao2026pisco,pan2026nova}.

Paired datasets support these advances across tasks. Se\~norita-2M, Goku, and OpenVE cover broad editing tasks, while EffectErase and ROSE provide object removal / insertion pairs~\citep{zi2025senorita,liang2026goku,he2025openve,fu2026effecterase,miao2025rose}. ALIVE focuses on first-frame-guided object insertion with coherent responses to source actions, supported by interaction-focused editing pairs and multi-level prompt annotations for studying interaction guidance.

\noindent\textbf{Interactive video editing.} Interactive editing spans user control in streaming workflows and physical interactions within scenes. EditStream, JoyAI-Video-Edit, StreamEdit, and Vidu S2-Editing support streaming editing~\citep{zhou2026editstream,xiao2026joyai,jiao2026streamedit,zhang2026vidu}. EgoPlay times edits via user-specified source events~\citep{mai2026egoplay}. EgoEdit supports instruction-guided egocentric editing and real-time streaming~\citep{li2026egoedit}. Within scenes, VOID revises downstream physical interactions after object removal~\citep{motamed2026void}, while DynaEdit edits actions and dynamics through text~\citep{kulikov2026versatile}. We study first-frame-guided object insertion, where added objects respond to source actions without user-provided interaction descriptions.

\noindent\textbf{VLM guidance for video editing.} UniVideo and Omni-Video 2 integrate multimodal understanding with video generation and editing~\citep{wei2026univideo,yang2026omni}. DynVFX uses a VLM to describe a scene augmented with dynamic content~\citep{yatim2025dynvfx}; Aurora plans edits and obtains missing conditions through tools~\citep{yu2026aurora}. VOID identifies regions affected by object removal to guide counterfactual generation~\citep{motamed2026void}. ALIVE trains a VLM to infer the inserted object's possible responses from the source video, edited first frame, and object-only instruction, expressing them as temporally localized prompts guiding corresponding video intervals.

\vspace{-2mm}
\section{ALIVE Dataset and Benchmark}
\label{headings}
\label{sec:dataset}
\vspace{-1mm}

Interaction-focused editing pairs are scarce, and evaluating object insertion requires assessing responses to source actions beyond appearance.
We introduce ALIVE, comprising editing pairs from four complementary sources, multi-level prompt annotations for studying interaction guidance, and the ALIVE-interaction benchmark for evaluating interaction quality and source preservation.

\subsection{Interaction Pair Construction}
\label{sec:data_construction}

\begin{figure}[!t]
    \centering
    \includegraphics[width=\linewidth]{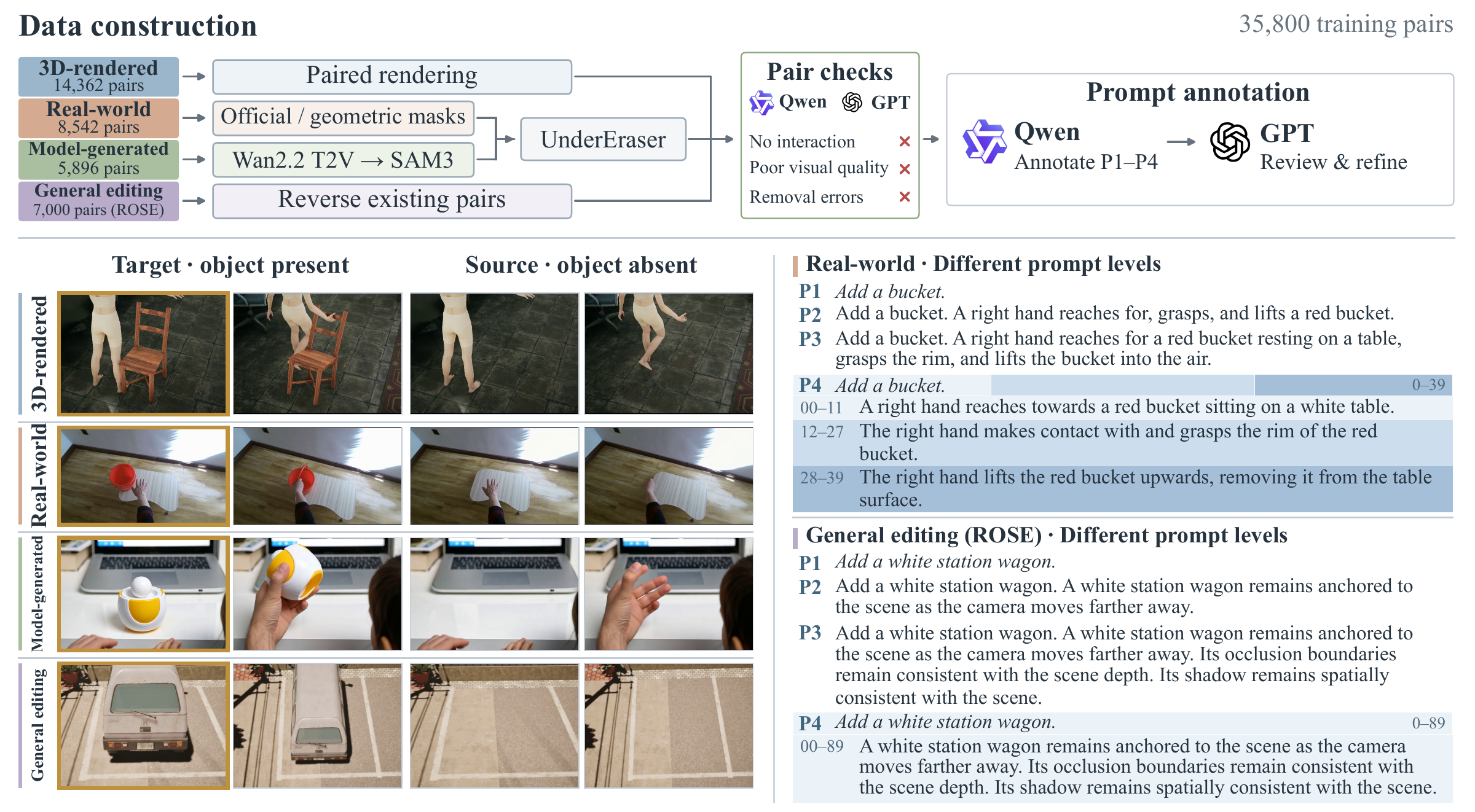}
    \vspace{-5mm}
    \caption{\textbf{ALIVE data construction and prompt annotation.} Top: construction and quality filtering.
    Bottom: editing pairs from four sources (left) and multi-level prompts (right). P0 (no text) and P1 (object identity only) omit interaction descriptions; P2--P4 provide interaction guidance. Gold outlines mark edited first frames; P4 shading marks chunks with inclusive frame ranges.}
    \label{fig:data_overview}
    \vspace{-3mm}
    \end{figure}

Constructing these pairs requires finding suitable interaction videos and varying the target object's presence while preserving the surrounding scene and action. We combine \textit{3D-rendered videos}, \textit{Model-generated videos}, and \textit{Real-world videos} for interaction examples across visual domains, supplemented by \textit{General editing pairs (ROSE)} for insertion without object interaction. Each pair comprises an object-absent source video $X$ and an object-present target video $Y$; its first frame provides the edited first frame $E_0$. Figure~\ref{fig:data_overview} summarizes construction.

\noindent\textbf{3D-rendered videos.} To obtain precisely aligned editing pairs, we use 3D assets from SpaceTimePilot~\citep{huang2025spacetimepilot,lu2025humoto}. We manually select segments featuring object interactions and render each scene from multiple first- and third-person viewpoints. For each viewpoint, we render sequences with and without the target object while keeping the surrounding scene, camera trajectory, and animation aligned. These renders directly provide $Y$, $X$, and object masks.

\noindent\textbf{Model-generated videos.} To broaden interaction and scene coverage beyond available assets, we use GPT to design prompts covering common interaction types and synthesize object-present videos with Wan2.2~\citep{wan2025wan}. SAM3~\citep{carion2026sam} segments and tracks the target object, and UnderEraser~\citep{liu2026understanding} removes it to construct the paired source video.

\noindent\textbf{Real-world videos.} For real-world visual supervision, we curate interaction clips from ARCTIC~\citep{fan2023arctic}, HOCap~\citep{wang2025hocap}, HOI4D~\citep{liu2022hoi4d}, and HOT3D~\citep{banerjee2025hot3d}. Target masks come from provided segmentations or projected object geometry, with refinement as needed. UnderEraser removes the target while retaining the surrounding action.

\noindent\textbf{General editing pairs (ROSE).} Objects should respond to interactions but also behave appropriately in their absence, for example by remaining stationary in the scene. We include general editing pairs from ROSE~\citep{miao2025rose} to support this behavior, alongside effects such as shadows and reflections. We reverse its removal pairs: the object-absent result becomes $X$, and the original object-present video becomes $Y$.

\noindent\textbf{Pair quality.} We use Qwen3.6-27B~\citep{qwen36_27b} and GPT-5.6 Luna/Sol for source-specific quality checks, excluding interaction videos with no visible object interaction or poor visual quality, and pairs with incomplete or visibly incorrect object removal.

Our 35,800 unique pairs comprise 14,362 3D-rendered, 5,896 model-generated, 8,542 real-world, and 7,000 ROSE general editing pairs. They provide precisely aligned pairs, common interactions, real-world examples, and cases without object interaction. Section~\ref{sec:exp_data} evaluates different sources.

\vspace{-1mm}
\subsection{Multi-Level Interaction Annotation}
\label{sec:data_annotation}

We investigate whether prompts improve object interactions in video editing and which descriptions are most effective. We annotate P1--P4 with varying semantic detail and temporal structure, alongside P0 (no text). P1 names only the added object. Neither P0 nor P1 describes interactions, leaving the editor to infer plausible responses from the source video and $E_0$. P2 describes the principal interaction; P3 details its sequence and outcome; P4 describes temporal chunks. P2--P4 provide interaction information to the editor. P4 can span the full clip when no phase decomposition is needed. Figure~\ref{fig:data_overview} illustrates these differences. These annotations enable VLM-predicted interaction prompts from the source video, $E_0$, and P1, without user-specified interaction descriptions (Section~\ref{sec:method_vlm}).

Qwen3.6-27B assists annotation using ordered target-video frames, target masks, and available object or action metadata, followed by selective GPT verification and refinement against the videos. Inconsistent descriptions are corrected or excluded at the affected level. For ROSE cases without object manipulation, prompts describe scene motion, occlusion, and visible object-related effects.

\vspace{-1mm}
\subsection{Evaluation Benchmarks}
\label{sec:data_benchmark}

\noindent\textbf{ALIVE-interaction benchmark.} We construct insertion tasks to assess coherent object responses to source actions. Each case provides an object-absent source video, an edited first frame, an object-present reference video, and P1--P4 annotations for prompt-level comparisons (Table~\ref{tab:prompt_guidance}), with P0 using no text. Its 128 cases include held-out samples from the 3D-rendered (10), model-generated (42), and real-world sources (19: ARCTIC 7, HOCap 1, HOI4D 5, and HOT3D 6), plus two out-of-domain subsets: 31 HOIGen-1M~\citep{liu2025hoigen} cases and 26 videos we recorded (RealShot). \textbf{General video object insertion benchmark.} Alongside interaction-focused editing, we evaluate general insertion on 103 cases: 60 from ROSE and 43 from EffectErase. A common protocol evaluates editing fidelity, source preservation, and visual coherence on both benchmarks (Section~\ref{sec:exp_setup}).

\vspace{-2mm}
\section{Method}
\label{others}
\label{sec:method}
\vspace{-1mm}

\begin{figure}[!t]
\centering
\includegraphics[width=\linewidth]{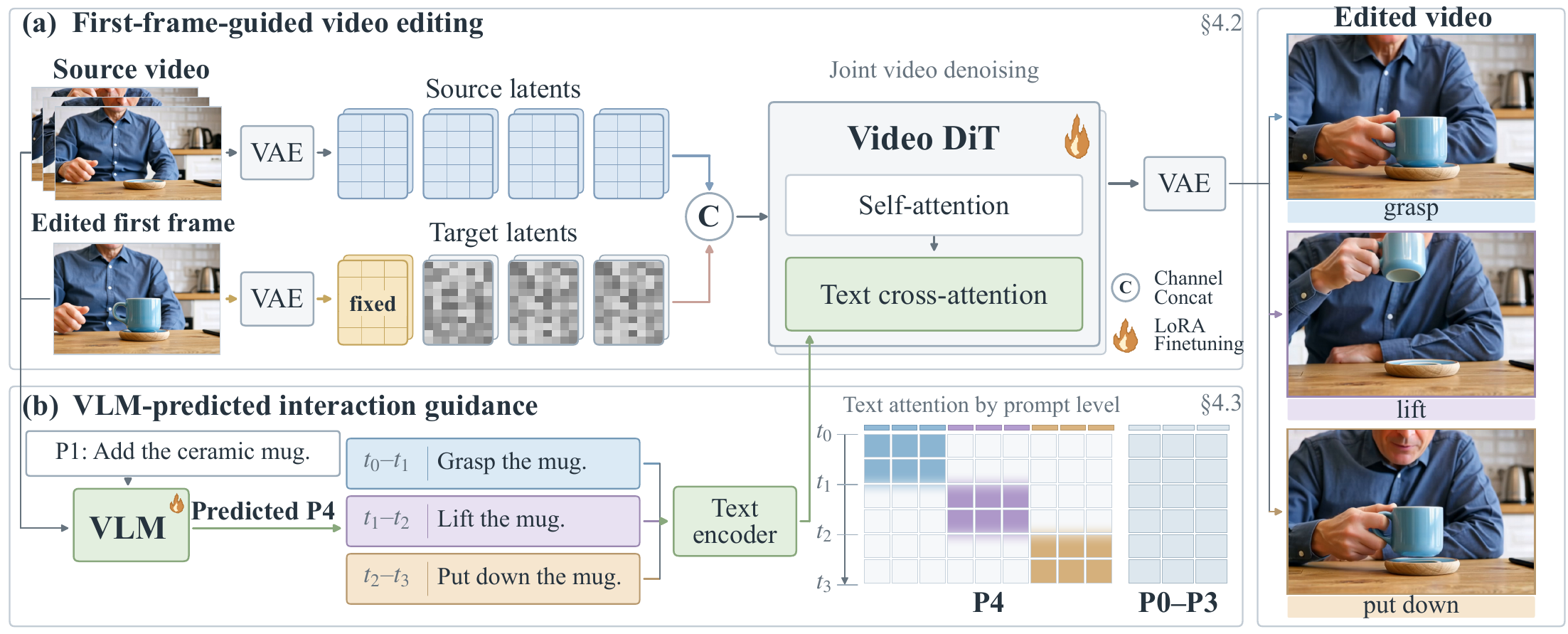}
\vspace{-3mm}
\caption{\textbf{ALIVE overview.} (a) Source and noisy target latents are channel-concatenated (C); the edited first-frame latent stays clean (gold). Flames mark editor and VLM LoRA fine-tuning. (b) A supervised VLM predicts P4 descriptions and intervals from the same visual inputs and P1. P4 texts are independently encoded. Attention maps contrast shared P0--P3 with time-local P4 conditioning. }
\label{fig:method_overview}
\vspace{-3mm}
\end{figure}

We present a first-frame-guided diffusion editor that learns inserted objects' responses to source actions (Figure~\ref{fig:method_overview}). Section~\ref{sec:method_editor} details its visual and prompt conditioning and training; Section~\ref{sec:method_vlm} describes how a VLM uses visual understanding to infer P4 chunk prompts from the same inputs for temporal guidance.

\subsection{Task Definition}
\label{sec:method_task}

Given a source video $X=(x_0,\ldots,x_{T-1})$, an edited first frame $E_0$, and an identity-only insertion instruction $p_1$, we aim to generate an edited video in which the added object participates coherently in the source action while unrelated scene content is preserved. The edited frame specifies the object's appearance and initial placement, $p_1$ names the object, and $X$ provides the action context. The editor $G_\theta$ infers subsequent object interactions from these inputs:
\begin{equation}
    \hat Y = G_\theta(X,E_0,p_1).
\end{equation}
The complete target video $Y$ provides training supervision but is unavailable at inference.

\subsection{Video Editing Model}
\label{sec:method_editor}

\noindent\textbf{Visual conditioning.} To propagate the initial insertion while preserving source actions, we condition the editor on $E_0$ and the full source video. Our main implementation adapts LTX-2.5~\citep{hacohen2026ltx}. A frozen video encoder maps source--target pairs to aligned latent grids $z_X$ and $z_Y$. We concatenate clean source latents with noisy target latents $z_t$ at matching spatiotemporal positions:
\begin{equation}
    h_t=W_{\mathrm{in}}[z_t\,;\,z_X]_{\mathrm{channel}}+b.
    \label{eq:source_conditioning}
\end{equation}
This supplies source context without increasing the number of video tokens. We initialize the target half of $W_{\mathrm{in}}$ from the pretrained projection and the source half to zero. The encoded $E_0$ anchors the first target frame; its conditioned tokens remain clean throughout denoising.

\noindent\textbf{Prompt conditioning.} The editor supports clip-level and temporally localized text conditioning. P0 uses an empty prompt while retaining the source video and $E_0$. For P1--P3, the backbone's standard text-conditioning path encodes the full prompt as one context shared across frames: P1 specifies object identity; P2 and P3 also describe interactions.

P4 aligns text guidance with source-action stages through descriptions $d_k$ and temporal intervals $I_k$. Each $d_k$ specifies the added object and its interaction during $I_k$. We independently encode these descriptions into contexts $C_k$ of length $L_k$ and concatenate them. For video token $i$, let $w_{ik}$ denote its normalized temporal overlap with $I_k$. Attention to text token $j$ in context $k$ is
\begin{equation}
    A_{ij}=\operatorname{softmax}_{j}
    \left(\frac{q_i^\top k_j}{\sqrt d}+\log\frac{w_{ik}}{L_k}\right),
    \qquad j\in C_k,
    \label{eq:chunk_attention}
\end{equation}
where $q_i$, $k_j$, and $d$ are the query, key, and key dimension. Zero overlap masks the context; dividing by $L_k$ prevents longer descriptions from gaining prior attention mass solely through length. This routing modifies only video-to-text cross-attention; all video chunks are denoised jointly.

\noindent\textbf{Training.} We train a single editor with a mixture of prompt levels P0--P4 using LoRA~\citep{hu2021lora} on ALIVE pairs. For each sampled level, we use pairs with valid annotations for that level. For unconditioned target tokens, the noisy latent at noise level $t$ is $z_t=(1-t)z_Y+t\epsilon$, with $\epsilon\sim\mathcal N(0,I)$. We optimize the native flow-matching objective~\citep{lipman2022flow},
\begin{equation}
    \mathcal L_{\mathrm{edit}}
    =\mathbb E_{Y,X,p,t,\epsilon}
    \bigl[\|v_\theta(z_t,t;z_X,E_0,p)-(\epsilon-z_Y)\|_{M}^{2}\bigr],
\end{equation}
where $v_\theta$ is the editor's velocity prediction with parameters $\theta$, $p$ is the sampled prompt, and $\|\cdot\|_M^2$ averages squared error over target tokens not conditioned on $E_0$. For VLM-guided editing, we condition the editor on P4 prompts predicted from the original inputs, as described next.

\vspace{-1mm}
\subsection{VLM-Based Interaction Guidance}
\label{sec:method_vlm}
\vspace{-1mm}

To guide object interactions over time, we use a VLM's visual understanding to infer P4 chunk prompts from the source video, edited first frame, and object-only instruction. These prompts describe plausible interactions and their temporal intervals without additional interaction description.

We LoRA-adapt Qwen3.6-27B on the P4 annotations in Section~\ref{sec:data_annotation}. Given temporally ordered source frames, $E_0$, and P1, denoted by $u=(X,E_0,p_1)$, the predictor learns chunk descriptions and intervals through cross-entropy on response tokens only. Annotation may use the target video, but prediction uses only $u$ during training and inference. At inference, the VLM $F_\phi$ predicts P4 guidance $\hat p_4$, replacing $p_1$ as the editor's text condition:
\begin{equation}
    \hat p_4=F_\phi(X,E_0,p_1), \qquad
    \hat Y=G_\theta(X,E_0,\hat p_4).
\end{equation}
The routing in Section~\ref{sec:method_editor} applies each description to its temporal interval, guiding joint generation of the complete video.

\vspace{-1mm}
\section{Experiments}
\label{sec:experiments}
\vspace{-1mm}
We evaluate whether ALIVE enables inserted objects to participate in source actions while retaining general insertion quality. We then examine how training-data composition supports these capabilities and whether interaction prompts provide further improvements.

\vspace{-1mm}
\subsection{Experimental Setup}
\vspace{-1mm}
\label{sec:exp_setup}

We evaluate on the two benchmarks introduced in Section~\ref{sec:data_benchmark}.

\noindent\textbf{Compared methods.} We compare AnyV2V, Se\~norita, NovaEdit, PropFly, and I2VEdit using their official conditioning interfaces and prompt formats. Both ALIVE variants share an LTX-2.5 model trained on our 35,800 pairs with mixed prompt levels: ALIVE uses object-identity text (P1), while ALIVE + VLM uses P4 predicted from the same source video, edited first frame, and P1. P2--P4 add interaction guidance beyond P1's object identity. Following their official prompt formats, AnyV2V and PropFly receive P2 descriptions, marked $\dagger$ for this additional guidance. We exclude instruction-only editors because text specifies initial object placement less precisely than $E_0$; placement differences may prevent the intended interaction, confounding comparisons against methods sharing the edited first frame. We also exclude mask-guided VACE and LoRA-Edit configurations because per-frame masks provide object states and motion that our task requires the editor to infer. See Appendix~\ref{sec:exp_details}--\ref{sec:supp_ablations} for training, inference, and additional ablations.

\noindent\textbf{Evaluation.} A reference-aware VLM judge (\texttt{gpt-5.6-sol}, reasoning effort \texttt{high}) assesses interaction fidelity, source preservation, and visual coherence under a shared rubric across both benchmarks. It receives 16 uniformly sampled output frames, including both endpoints, with corresponding source and reference frames, $E_0$, and the object-addition instruction. For I2VEdit, we evaluate all 14 output frames from its official implementation.
Overall aggregates six dimensions on a 0--100 scale; Table~\ref{tab:main_comparison} reports Overall, task fidelity, and motion/propagation. Beyond VLM scores, we follow EditVerse~\citep{ju2026editverse}, reporting PickScore, frame-level CLIP and video-level ViCLIP alignment, and CLIP/DINOv2-based temporal consistency. Instructions, preprocessing alignment, and scoring details appear in Appendix~\ref{sec:eval_protocol}.

\begin{table}[!t]
\centering
\small
\setlength{\tabcolsep}{2pt}
\renewcommand{\arraystretch}{1.}
\caption{Comparison on interaction-focused and general object insertion. Both ALIVE variants share the LTX-2.5 (22B) editor; VLM P4 is predicted from source/$E_0$/P1. $\dagger$ marks interaction descriptions supplied. VLM scores use a 0--100 scale; automatic metrics retain native scales. Higher is better throughout; bold/underline denote best/second-best scores, including ties.}
\label{tab:main_comparison}
\scalebox{0.95}{%
\begin{tabular*}{\linewidth}{@{\extracolsep{\fill}}llrrrrrrrr@{}}
\toprule
 & & \multicolumn{3}{c}{VLM evaluation} & \multicolumn{5}{c}{Automatic metrics} \\
\cmidrule(lr){3-5}\cmidrule(lr){6-10}
Method & Guidance & Overall & Task & Motion & PickScore & \shortstack{Frame\\CLIP} & \shortstack{Video\\ViCLIP} & TC-CLIP & TC-DINO \\
\midrule
\multicolumn{10}{l}{\textit{ALIVE-interaction benchmark (128 cases)}} \\
\midrule
Se\~norita & P1 & 60.24 & 53.52 & 36.52 & 19.95 & 24.04 & 19.09 & 0.9572 & 0.9494 \\
NovaEdit & P0 & 36.36 & 8.59 & 0.39 & 18.94 & 21.15 & 15.41 & 0.9744 & 0.9734 \\
I2VEdit & P0 & 56.55 & 48.44 & 33.79 & 19.68 & 23.48 & 18.61 & 0.9363 & 0.9132 \\
AnyV2V & P2$^\dagger$ & 42.75 & 44.53 & 31.25 & 19.77 & \underline{24.12} & 19.48 & 0.9141 & 0.8916 \\
PropFly & P2$^\dagger$ & 42.31 & 35.94 & 21.68 & 19.40 & 22.81 & 17.58 & 0.9390 & 0.9274 \\
\midrule
ALIVE & P1 & \underline{86.71} & \underline{93.75} & \underline{86.13} & \textbf{20.34} & \textbf{25.00} & \underline{20.18} & \textbf{0.9795} & \textbf{0.9759} \\
ALIVE + VLM & VLM P4 & \textbf{87.66} & \textbf{93.95} & \textbf{86.91} & \underline{20.33} & \textbf{25.00} & \textbf{20.22} & \underline{0.9791} & \underline{0.9753} \\
\midrule
\multicolumn{10}{l}{\textit{general video object insertion benchmark (103 cases)}} \\
\midrule
Se\~norita & P1 & 88.39 & 88.83 & 83.74 & 20.30 & \textbf{23.98} & 19.35 & 0.9713 & 0.9723 \\
NovaEdit & P0 & 45.66 & 27.18 & 19.66 & 19.49 & 21.49 & 16.74 & 0.9820 & 0.9813 \\
I2VEdit & P0 & 82.21 & 84.95 & 79.13 & 20.09 & 23.50 & 19.38 & 0.9556 & 0.9403 \\
AnyV2V & P2$^\dagger$ & 48.03 & 56.07 & 47.09 & 20.09 & 23.71 & \textbf{20.15} & 0.9220 & 0.9020 \\
PropFly & P2$^\dagger$ & 64.02 & 64.32 & 55.58 & 19.93 & 22.88 & 18.62 & 0.9603 & 0.9545 \\
\midrule
ALIVE & P1 & \textbf{92.28} & \textbf{94.17} & \textbf{89.32} & \underline{20.37} & 23.74 & 19.47 & \underline{0.9845} & \underline{0.9835} \\
ALIVE + VLM & VLM P4 & \underline{91.71} & \underline{93.20} & \underline{88.35} & \textbf{20.38} & \underline{23.85} & \underline{19.63} & \textbf{0.9849} & \textbf{0.9849} \\
\bottomrule
\end{tabular*}
}
\vspace{-3mm}
\end{table}

\vspace{-2mm}
\subsection{Comparison with Existing Video Editors}
\label{sec:exp_main}

ALIVE substantially improves interaction-focused insertion while retaining strong general insertion quality (Table~\ref{tab:main_comparison}). With object-identity text alone, it exceeds the strongest evaluated baseline by 26.46 points on the ALIVE-interaction benchmark (86.71 vs.\ 60.24) and by 3.89 points on general insertion (92.28 vs.\ 88.39). Its motion/propagation score on the interaction benchmark rises to 86.13, compared with 36.52 for Se\~norita, supporting more coherent coordination between inserted objects and source actions. VLM-predicted P4 further raises the interaction Overall score to 87.66. For general insertion, P1 remains stronger than predicted P4 (92.28 vs.\ 91.71), although both outperform the evaluated baselines. Predicted interaction guidance therefore builds on an already capable editor. Section~\ref{sec:exp_guidance} examines this contribution in detail.

\begin{figure}[!tb]
    \centering
    \includegraphics[width=\linewidth]{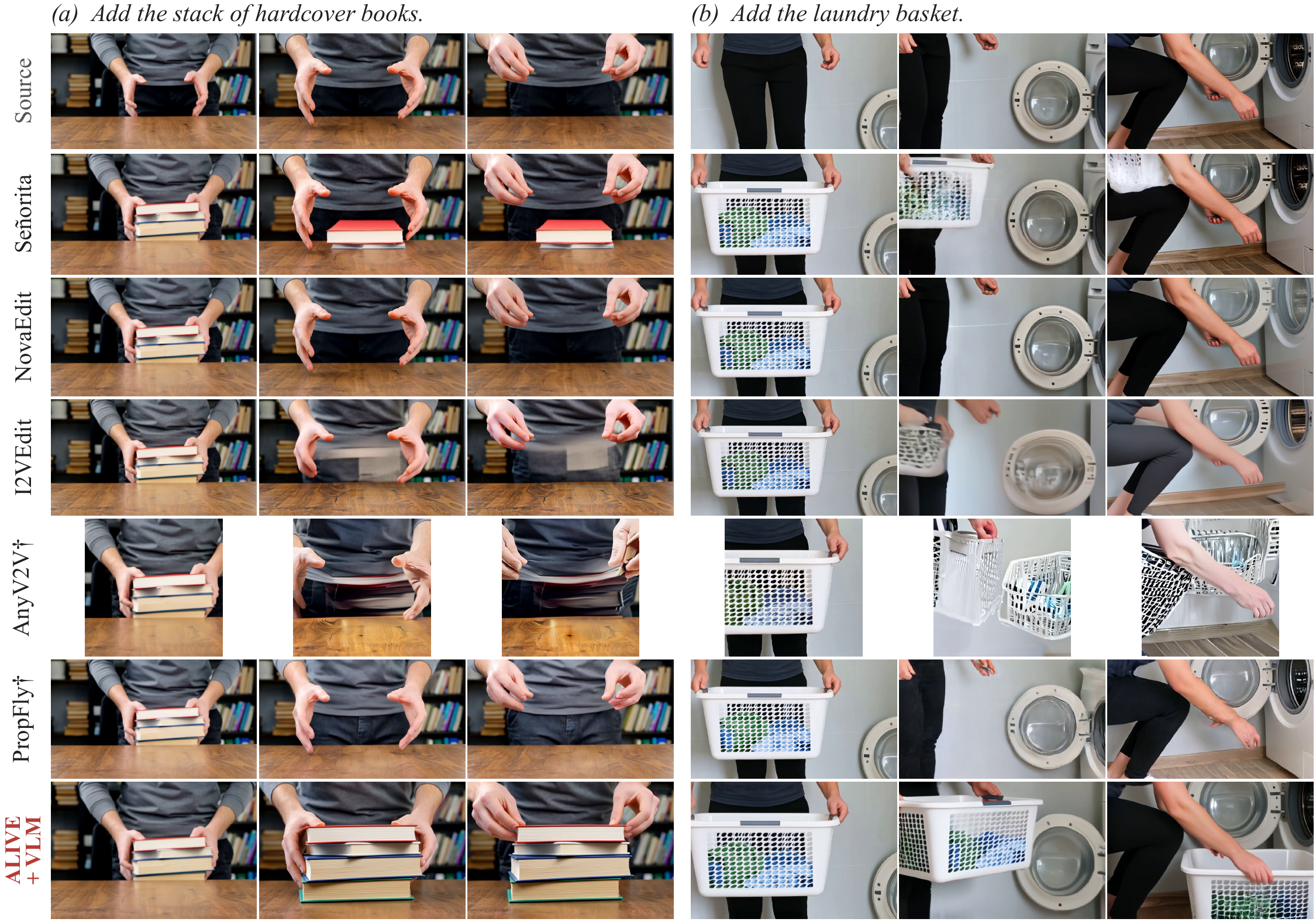}
    \caption{\textbf{Interaction-focused object insertion.} Book handling (a) and basket carrying (b), each shown at the first frame, 40\%, and 90\% of the video. Square AnyV2V frames are displayed more narrowly. $\dagger$ marks provided interaction guidance, following the methods' official prompt formats.}
    \label{fig:qualitative}
    \vspace{-3mm}
\end{figure}

Figure~\ref{fig:qualitative} compares book handling and basket carrying samples. ALIVE + VLM moves the inserted objects with the hands. In these examples, baselines can leave objects stationary, omit them, or distort their shape.

\vspace{-1mm}
\subsection{Ablation Studies}
\label{sec:exp_ablations}

\vspace{-1mm}
\subsubsection{Data Sources}
\label{sec:exp_data}

We examine how each data source contributes to interaction-focused and general object insertion, and whether combining them benefits both. We compare single-source training with a four-source mixture sampled in proportion to source size. All conditions use LTX-2.5, rank-128 LoRA, P1, and 5k training steps, with fixed evaluation cases and generation seeds. Table~\ref{tab:source_mixture} lists source-pool sizes.

\begin{table}[!htb]
\centering
\small
\setlength{\tabcolsep}{2pt}
\renewcommand{\arraystretch}{1.12}
\caption{Single-source and mixed-source training with the same 5k-step budget. All scores use our 16-frame evaluation (0--100; $\uparrow$). Pooled averages all 231 cases with equal weight: $(128\,S_{\mathrm{int}}+103\,S_{\mathrm{gen}})/231$, where $S_{\mathrm{int}}$ and $S_{\mathrm{gen}}$ are the benchmark means. Bold and underline mark the best and second-best scores per column.}
\label{tab:source_mixture}
\begin{tabular*}{\linewidth}{@{\extracolsep{\fill}}lrrrrrrrr@{}}
\toprule
Training source & Pairs & Pooled & \multicolumn{3}{c}{\shortstack{ALIVE-interaction\\benchmark\\($N=128$)}} & \multicolumn{3}{c}{\shortstack{general video object\\insertion benchmark\\($N=103$)}} \\
\cmidrule(lr){4-6}\cmidrule(lr){7-9}
 & & Overall & Overall & Task & Motion & Overall & Task & Motion \\
\midrule
3D-rendered videos & 14,362 & 74.85 & 73.08 & 76.37 & 62.89 & 77.06 & 78.16 & 71.12 \\
Real-world videos & 8,542 & \underline{81.51} & 80.19 & 88.48 & 79.49 & 83.16 & 83.50 & 80.10 \\
Model-generated videos & 5,896 & 81.26 & \textbf{82.81} & \textbf{90.82} & \textbf{82.23} & 79.32 & 79.85 & 71.84 \\
General editing pairs (ROSE) & 7,000 & 70.93 & 60.12 & 60.74 & 45.12 & \underline{84.36} & \underline{88.35} & \underline{81.31} \\
All four sources & 35,800 & \textbf{84.39} & \underline{82.35} & \underline{89.65} & \underline{82.03} & \textbf{86.93} & \textbf{88.83} & \textbf{83.01} \\
\bottomrule
\end{tabular*}
\vspace{-3mm}
\end{table}

Different sources favor different capabilities: model-generated pairs perform best on interaction, while ROSE leads individual sources on general insertion. However, ROSE-only training scores just 60.12 on the ALIVE-interaction benchmark, compared with 82.35 for the mixture, highlighting the value of interaction-focused training pairs. The mixture remains within 0.46 points of model-generated-only training on interaction while improving general insertion by 7.61 points. Its highest general-insertion and pooled scores support the complementary value of our data sources.

\vspace{-1mm}
\subsubsection{Prompt Levels and VLM Guidance}
\label{sec:exp_guidance}

Having established the editor's interaction capability, we test whether explicit descriptions of object behavior provide useful additional guidance. Table~\ref{tab:prompt_guidance} compares annotated and VLM-predicted prompts while keeping editor weights, visual inputs, and generation seeds fixed. P0 supplies no text, P1 names the object, and annotated P2--P4 additionally describe interactions. VLM variants predict those descriptions from source/$E_0$/P1.

\begin{figure}[!tb]
\centering
\includegraphics[width=\linewidth]{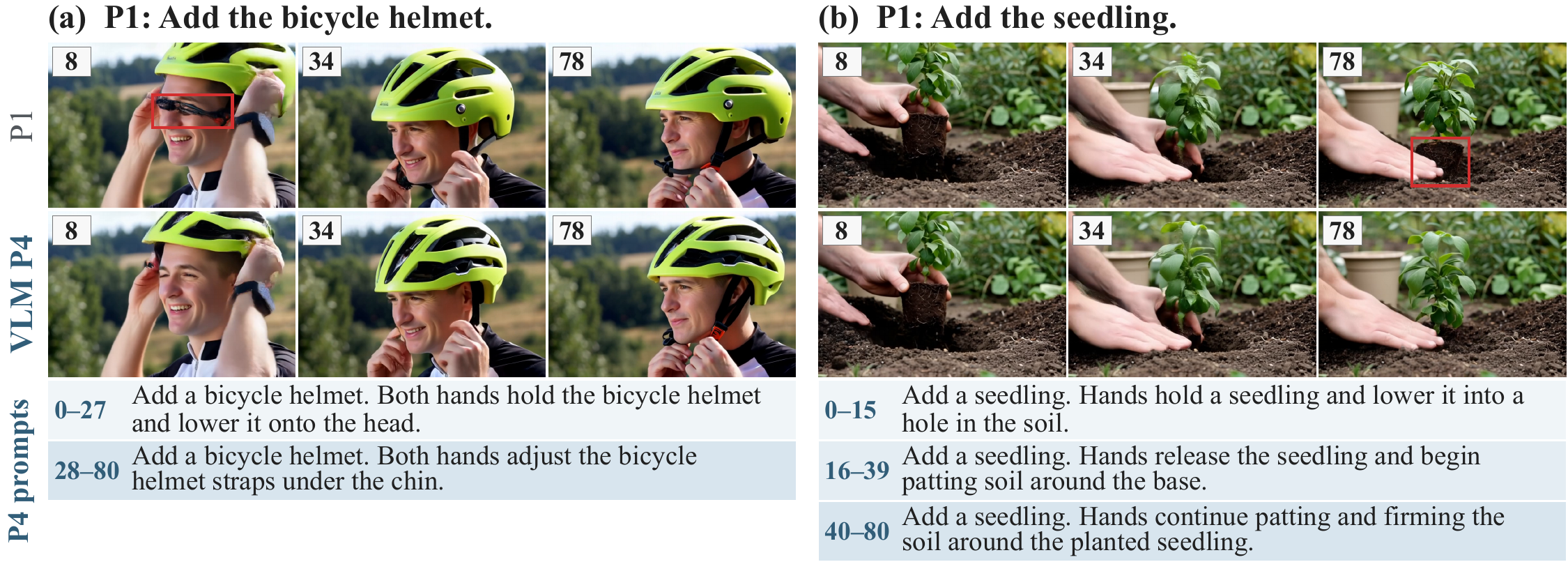}
\caption{\textbf{Effect of VLM-predicted interaction guidance.} P1 and predicted P4 condition the same editor with fixed visual inputs. Predicted prompts and inclusive frame ranges appear below. Red boxes highlight strap artifacts around the eyes (a) and an exposed root ball (b) in the P1 results.}
\label{fig:vlm_guidance}
\vspace{-3mm}
\end{figure}

\noindent\textbf{Effects of prompt levels.} P0 and P1 achieve Overall scores of 85.38 and 86.71, respectively, showing that the editor can infer object interactions from visual inputs.
P2 provides little additional benefit, while the more detailed P3 and temporally localized P4 improve Overall to 87.37 and 88.69. These results suggest that the form of interaction guidance matters, with P4 performing best among the annotated formats.

\noindent\textbf{VLM-predicted guidance.} Predicted P4 achieves the highest Overall among the VLM variants, raising P1 from 86.71 to 87.66 using the original editing inputs. P4's gains include temporal consistency and background preservation as well as motion/propagation. Thus, predicted P4 offers an improvement in overall interaction-editing quality.

Figure~\ref{fig:vlm_guidance} connects the predicted descriptions to the resulting object behavior. P4 specifies ``lower it onto the head'' before adjusting the helmet straps under the chin, clarifying the placement sequence; its output avoids the strap artifacts around the eyes seen with P1. For the seedling, it describes lowering, releasing, and firming the surrounding soil; the generated plant is better seated in the soil as the hands press around it.

\begin{table}[!t]
\centering
\small
\setlength{\tabcolsep}{3pt}
\renewcommand{\arraystretch}{1.12}
\caption{Prompt guidance on the ALIVE-interaction benchmark. All rows use the same editor, visual inputs, and case-specific seeds. Annotated P2--P4 supply interaction information; VLM variants infer it from source/$E_0$/P1. P4 uses temporal chunks. Scores are 0--100 ($\uparrow$); bold and underline mark first and second place. ID: identity; BG: background; Temp.: temporal consistency.}
\label{tab:prompt_guidance}
\begin{tabular}{@{}lrrrrrrrr@{}}
\toprule
Guidance & $N$ & Task & ID & Motion & BG & Temp. & Visual & Overall \\
\midrule
P0 & 128 & 91.41 & 81.84 & 83.40 & \underline{93.36} & 80.47 & 75.00 & 85.38 \\
P1 & 128 & 93.75 & 83.98 & 86.13 & 91.80 & 81.05 & 75.20 & 86.71 \\
P2 & 128 & 93.16 & 82.23 & 86.91 & 92.38 & 80.66 & 77.15 & 86.68 \\
P3 & 128 & \underline{94.73} & 83.40 & \underline{88.48} & 91.60 & 81.25 & 75.59 & 87.37 \\
P4 & 128 & \textbf{95.12} & \textbf{85.94} & \textbf{88.67} & \textbf{93.55} & \textbf{82.81} & \textbf{78.32} & \textbf{88.69} \\
\midrule
\multicolumn{9}{l}{\textit{VLM-enhanced P1}} \\
VLM P2 & 128 & 93.55 & \underline{84.57} & 87.89 & 91.80 & \underline{81.84} & 76.37 & 87.33 \\
VLM P3 & 128 & 94.34 & 83.20 & 86.33 & 92.19 & \underline{81.84} & \underline{77.34} & 87.17 \\
VLM P4 & 128 & 93.95 & \underline{84.57} & 86.91 & \underline{93.36} & \textbf{82.81} & 76.76 & \underline{87.66} \\
\bottomrule
\end{tabular}
\end{table}

\noindent\textbf{Limitations and future work.} Despite these improvements, ALIVE faces limitations. The source video and edited first frame do not fully specify geometry, contact conditions, or forces, leaving multiple plausible object responses. For example, a soft object may deform differently under pressure depending on stiffness, while a partially occluded grasp may allow a cup to remain upright or tilt during lifting. ALIVE aims to generate plausible object responses consistent with source actions, without implausible deformation, interpenetration, or unsupported floating. Additionally, VLM-predicted guidance errors can cause incorrect behavior, and the editor still struggles with complex cloth deformations and rapid, large-angle rotations. Future work includes improving guidance reliability, strengthening the editor's ability to generate complex interactions, and incorporating explicit physical information for greater accuracy and control. Combining ALIVE with Self Forcing~\citep{huang2025selfforcing} or EditStream~\citep{zhou2026editstream} could enable real-time streaming object insertion.

\section{Conclusion}
We presented ALIVE, a framework for inserting objects that participate coherently in a source video's interactions. Our dataset combines complementary construction routes, and our benchmarks assess both interaction-focused and general object insertion. Experiments show the benefit of interaction-focused editor training and clarify the effects of data composition and prompt guidance. VLM-predicted guidance further improves interaction editing without additional user input, with P4 performing best among the evaluated prediction formats.

\FloatBarrier
\section*{Acknowledgments}
The implementation and experiments reported in this paper were carried out by Zhenghong Zhou at the University of Rochester. This work was partially supported by the university's Goergen Institute for Data Science and Artificial Intelligence. We gratefully acknowledge use of the research computing resources of the Empire AI Consortium, Inc.~\citep{Bloom2025EmpireAI}, with support from Empire State Development of the State of New York, the Simons Foundation, and the Secunda Family Foundation.

\begingroup
\catcode`\&=12
\bibliography{references}
\endgroup
\bibliographystyle{plainnat}

\FloatBarrier
\clearpage
\appendix
\section{Appendix}
\label{sec:supplement}

\subsection{Training Details}
\label{sec:exp_details}

\noindent\textbf{Diffusion editor training.} We adapt LTX-2.5 with rank-128 LoRA on 35,800 pairs, using 81-frame clips at $832\times480$. Training uses AdamW with an initial learning rate of $10^{-4}$ and linear decay, a global batch size of 16, and 20,000 updates. We use 16 GPUs per run (A100 80GB, H100, or H200), with one sample per GPU. Training first selects P0--P4 with probabilities 0.1/0.3/0.2/0.2/0.2, then samples a video pair with the corresponding prompt annotation (no text for P0). Data-source ablations use P1 and 5,000 updates with the same batch size, evaluation cases, and generation seeds. Both ALIVE variants share the main editor; at inference, they generate 81 frames at $768\times512$ with 30 denoising steps.

\noindent\textbf{VLM training.} We adapt Qwen3.6-27B with rank-64 LoRA ($\alpha=128$, dropout 0.05), training separate predictors for P2, P3, and P4. Each uses 16 A100 80GB GPUs, one sample per GPU, and two-step gradient accumulation, giving a global batch size of 32. We use AdamW with learning rate $5\times10^{-5}$, weight decay 0.01, and 32 warm-up updates followed by a constant learning rate. Inputs comprise 21 uniformly sampled source frames, $E_0$, and P1. We use the 2,048-update checkpoint for each predictor.

\subsection{Inference Settings and Efficiency}
\label{sec:inference_efficiency}

Table~\ref{tab:inference_efficiency} profiles the same six videos (three per benchmark) using one A100 80GB per video, with batch size one and a separate warm-up video. We report mean wall-clock time and the largest peak allocated GPU memory across the six cases. Timing includes conditioning, required inversion or per-video adaptation, sampling, decoding, and export; offline data preparation and one-time setup are excluded. Costs reflect each method's native output configuration and the memory-management settings specified below.

\begin{table}[!htbp]
\centering
\small
\setlength{\tabcolsep}{4pt}
\renewcommand{\arraystretch}{1.08}
\caption{Measured inference cost on A100 80GB. Output is width $\times$ height $\times$ frames; only denoising iterations are counted under Steps. Time includes each editor's required inversion and adaptation. Peak denotes PyTorch peak allocated memory in GiB. VLM prompt prediction is measured separately in the last row; dashes mean not applicable.}
\label{tab:inference_efficiency}
\begin{tabular}{@{}llrrr@{}}
\toprule
Method & Output & Steps & Time (s) & Peak (GiB) \\
\midrule
Se\~norita & $768\!\times\!448\!\times\!33$ & 30 & 109.18 & 28.04 \\
NovaEdit & $832\!\times\!480\!\times\!81$ & 50 & 407.70 & 19.63 \\
I2VEdit & $1024\!\times\!576\!\times\!14$ & 25 & 863.62 & 65.16 \\
AnyV2V & $512\!\times\!512\!\times\!16$ & 50 & 172.79 & 30.31 \\
PropFly & $832\!\times\!480\!\times\!25$ & 50 & 71.63 & 21.14 \\
\midrule
Wan 14B, P1 & $768\!\times\!512\!\times\!81$ & 30 & 304.64 & 64.46 \\
LTX 22B, P1 & $768\!\times\!512\!\times\!81$ & 30 & 154.16 & 39.43 \\
LTX 22B, predicted P4 & $768\!\times\!512\!\times\!81$ & 30 & 184.50 & 39.43 \\
\midrule
VLM P4 predictor & -- & -- & 22.75 & 52.27 \\
\bottomrule
\end{tabular}
\end{table}

AnyV2V uses 500 inversion steps; I2VEdit includes inversion, 250 motion-LoRA updates, and native per-video model loading. Se\~norita and PropFly run without CPU model offloading. Wan retains both diffusion experts on GPU while staging its text encoder and VAE; LTX uses its native no-offload setting. The P4 editor uses the fixed predicted prompts from the quality evaluation. Separately, the 2,048-update VLM is profiled on the same editing inputs. Summing these separately measured stages gives an estimated 207.25\,s per edit, excluding the overhead of switching models.

\subsection{Additional Ablations}
\label{sec:supp_ablations}

\subsubsection{Training Length}
\label{sec:exp_length}

We compare independently trained P1-only LTX-2.5 editors with 5k and 20k training steps on the same 35,800 pairs (Table~\ref{tab:training_length}). Evaluation cases and generation seeds are fixed, following the protocol in Section~\ref{sec:exp_setup}. The longer budget raises interaction Overall from 82.35 to 86.46 and general-insertion Overall from 86.93 to 89.20.

\begin{table}[!htbp]
\centering
\small
\setlength{\tabcolsep}{3pt}
\renewcommand{\arraystretch}{1.12}
\caption{Effect of training budget on P1-only LTX-2.5 editors (LoRA rank 128, batch 16). Scores report Overall ($\uparrow$) on both benchmarks.}
\label{tab:training_length}
\begin{tabular}{@{}lrrr@{}}
\toprule
Optimizer steps & Pairs & \shortstack{ALIVE-interaction\\benchmark} & \shortstack{general video object\\insertion benchmark} \\\midrule
5k & 35,800 & \underline{82.35} & \underline{86.93} \\
20k & 35,800 & \textbf{86.46} & \textbf{89.20} \\
\bottomrule
\end{tabular}
\end{table}

\subsubsection{Base Models and Model Size}
\label{sec:exp_capacity}

We compare Wan and LTX editors trained exclusively with P1 for 5k steps on the same 35,800 pairs (Table~\ref{tab:backbones}). Both benchmarks use the evaluation protocol in Section~\ref{sec:exp_setup}.

\begin{table}[!htbp]
\centering
\small
\setlength{\tabcolsep}{3pt}
\renewcommand{\arraystretch}{1.12}
\caption{Backbone comparison after 5k steps of P1-only training. Scores report Overall ($\uparrow$) on both benchmarks; bold and underline mark the best and second-best scores.}
\label{tab:backbones}
\begin{tabular}{@{}llrr@{}}
\toprule
Backbone & Steps & \shortstack{ALIVE-interaction\\benchmark} & \shortstack{general video object\\insertion benchmark} \\\midrule
Wan 2.2 TI2V (5B) & 5k & 71.85 & 81.54 \\
Wan 2.2 I2V (14B) & 5k & \textbf{89.82} & \textbf{89.44} \\
LTX-2.5 (22B) & 5k & \underline{82.35} & \underline{86.93} \\
\bottomrule
\end{tabular}
\end{table}

\noindent\textbf{Training efficiency.} For 81-frame clips at $832\times480$, LoRA rank 128, and batch 16 per model or expert, LTX-2.5 takes 4.95\,s per update on 16 A100 80GB GPUs; its 5k-step run takes 8.1 hours including checkpoint writes. Wan 14B takes approximately 45.4\,s per expert update on 32 A100 80GB GPUs (16 per noise expert), implying approximately 63 hours for 5k updates at this rate. Update times are steady-state medians; the Wan duration is an extrapolation, excluding setup, checkpoint overhead, and interruptions. LTX's stronger VAE compression ($8\times32\times32$ versus $4\times8\times8$ in time, height, and width) reduces the video-token sequence length. This supports its practical efficiency. Wan 14B achieves higher scores in the 5k-step comparison, while LTX-2.5 trains substantially faster and serves as our main experimental backbone.

\subsection{Evaluation Protocol}
\label{sec:eval_protocol}

We assess interaction fidelity, source preservation, and visual coherence using the VLM judge introduced in Section~\ref{sec:exp_setup}. Comparisons within each table use matched cases and generation seeds. Each candidate is evaluated independently with the same rubric and dimension weights across both benchmarks and all prompt levels. We reuse judgments for identical videos and evaluation inputs, retrying only failed requests without resampling successful scores.

\noindent\textbf{Evaluation inputs.} For each case, the judge receives source, reference, and candidate frames, the edited first frame $E_0$, and an object-addition instruction shared across methods (Table~\ref{tab:eval_inputs}). The ALIVE-interaction benchmark uses P1; the general video object insertion benchmark uses ``Add the object or person shown in the edited first frame.'' Reference frames beyond $E_0$ are provided only to the judge, not the editor.

\begin{table}[!htbp]
\centering
\small
\setlength{\tabcolsep}{5pt}
\renewcommand{\arraystretch}{1.12}
\caption{Inputs to the evaluation judge. The same roles and presentation are used for all benchmarks and methods. Only CANDIDATE is scored.}
\label{tab:eval_inputs}
\begin{tabular}{@{}p{0.20\linewidth}p{0.74\linewidth}@{}}
\toprule
Input & Role \\
\midrule
Basic instruction & Identify the intended addition without prescribing an action sequence. \\
SOURCE & Establish non-target scene content, camera motion, and interaction context. \\
Edited first frame $E_0$ & Specify the intended object or person and initial edit. \\
REFERENCE & Provide identity, contact, occlusion, interaction, and natural-effects evidence. \\
CANDIDATE & Generated result to evaluate at every displayed timepoint. \\
\bottomrule
\end{tabular}
\end{table}

\noindent\textbf{Frame sampling and alignment.} We align source and reference evidence to each generator's input preprocessing so that evaluation reflects the field of view and time span available during generation. We uniformly sample 16 native output frames, including both endpoints, and map them to the source frames selected by the generator. For I2VEdit, we use all 14 native frames without repetition or interpolation. Spatial alignment follows the generator's documented aspect-preserving cover resize and fixed center crop, including interpolation and rounding conventions.

At each timepoint, SOURCE, REFERENCE, and CANDIDATE appear side by side with their aspect ratios preserved. Four contact sheets show 5, 5, 5, and 1 timepoints; I2VEdit uses three sheets with 5, 5, and 4. The aligned $E_0$ is supplied separately as a PNG. Each labeled cell measures $384\!\times\!240$ pixels, and sheets use JPEG quality 88. Time labels indicate normalized positions rather than seconds. Layout margins, labels, and gaps are excluded from scoring; artifacts within video regions remain scoreable.

\noindent\textbf{Scoring and aggregation.} Table~\ref{tab:eval_dimensions} defines six dimensions and their weights. For candidate $i$ and dimension $d$, the judge assigns an integer $s_{i,d}\in\{0,1,2,3,4\}$: 0 denotes failure or contradictory evidence, 2 partial correctness with a material problem, and 4 full satisfaction supported by visible evidence. Scores 1 and 3 fall between these anchors. We assess plausible interactions consistent with SOURCE and $E_0$, allowing reasonable spatial and temporal differences from the reference (Figure~\ref{fig:eval_prompt}). Missing required interactions, broken contact, impossible motion, and source corruption remain errors.

\begin{table}[!htbp]
\centering
\small
\setlength{\tabcolsep}{5pt}
\renewcommand{\arraystretch}{1.12}
\caption{Evaluation dimensions. Weights are identical across benchmarks and ablations.}
\label{tab:eval_dimensions}
\begin{tabular}{@{}p{0.22\linewidth}p{0.62\linewidth}r@{}}
\toprule
Dimension & Criterion & Weight \\
\midrule
Task fidelity & Complete addition and coherent participation in interactions supported by SOURCE and $E_0$. & 0.25 \\
Object identity & Stable identity, geometry, and attributes consistent with $E_0$. & 0.15 \\
Motion / propagation & Plausible motion or appropriate stability, contact, timing, and occlusion. & 0.20 \\
Background preservation & Faithful non-target source content and camera motion. & 0.15 \\
Temporal consistency & Stable appearance and smooth transitions without flicker or disappearance. & 0.15 \\
Visual quality & Realistic boundaries, lighting, and natural effects. & 0.10 \\
\bottomrule
\end{tabular}
\end{table}

We aggregate scores as
\begin{equation}
\mathrm{Overall}_i=25\sum_{d=1}^{6}w_d s_{i,d},\qquad
\mathrm{Overall}=\frac{1}{N}\sum_{i=1}^{N}\mathrm{Overall}_i,\qquad
\sum_d w_d=1.
\end{equation}
Here, $N$ is the number of evaluated cases and $w_d$ the weight of dimension $d$. Dimension scores are also multiplied by 25 for reporting. The judge returns structured scores and explanations; its confidence indicates uncertainty but does not affect aggregation.

\noindent\textbf{Complete judge prompt.} Figure~\ref{fig:eval_prompt} reproduces the full task prompt supplied with the visual evidence. The placeholder \texttt{\textless basic instruction\textgreater} is replaced by the case's P1 instruction on the ALIVE-interaction benchmark, or ``Add the object or person shown in the edited first frame.'' on the general video object insertion benchmark. For I2VEdit, only ``The first four images show sixteen normalized-time samples'' changes to ``The first three images show fourteen normalized-time samples''; all scoring instructions remain identical.

\begin{samepage}
{\raggedright
\noindent\textbf{Structured response.} The request also supplies a JSON schema requiring five fields: \texttt{scores}, \texttt{justifications}, \texttt{failure\_flags}, \texttt{judge\_confidence}, and \texttt{summary}. Both \texttt{scores} and \texttt{justifications} contain all six dimension keys in Figure~\ref{fig:eval_prompt}; values are integers in $[0,4]$ and nonempty strings, respectively. Confidence is a number in $[0,1]$, and the summary is a nonempty string. Additional keys are disallowed in the root object and both dimension objects. Failure flags form a possibly empty array drawn from \texttt{missing\_target}, \texttt{wrong\_target}, \texttt{motion\_not\_followed}, \texttt{incomplete\_propagation}, \texttt{wrong\_motion\_or\_placement}, \texttt{source\_corruption}, \texttt{temporal\_breakdown}, \texttt{first\_frame\_mismatch}, and \texttt{catastrophic\_artifacts}.\par}
\end{samepage}

\begin{samepage}
\noindent\textbf{Additional metrics.} We compute PickScore~\citep{kirstain2023pick}, Frame CLIP~\citep{radford2021learning}, Video ViCLIP~\citep{wang2024internvid}, TC-CLIP, and TC-DINO (DINOv2~\citep{oquab2023dinov2}) on the same candidate videos using the official EditVerse implementations~\citep{ju2026editverse} and fixed captions. These metrics retain their native scales and do not enter the VLM Overall score. \par
\end{samepage}

\begin{figure}[!htbp]
\centering
\fbox{\begin{minipage}{\dimexpr\linewidth-2\fboxsep-2\fboxrule\relax}
\small\raggedright
\setlength{\parindent}{0pt}
\setlength{\parskip}{3pt}
Judge one anonymous candidate video. Do not use tools, inspect files, or infer a model identity. The first four images show sixteen normalized-time samples, with SOURCE, REFERENCE, and CANDIDATE columns. The final image is the actual edited first-frame condition E0. Score only CANDIDATE. SOURCE shows the scene and its motion before the addition. E0 specifies the intended added object or person. REFERENCE provides evidence of identity, interaction, contact, occlusion, and natural effects, not an exact trajectory or pixel-level target. Model names, expanded generator prompts, and annotated action phases are hidden. The same basic addition instruction is used for all methods on a case.

Judge whether the added object belongs coherently in the source scene and responds plausibly to its visible motion and interactions. Reasonable differences in position, path, pose, or timing from REFERENCE are acceptable when consistent with SOURCE and E0. Do not demand a specific action sequence that cannot be inferred from SOURCE and E0. This does not excuse missing interactions, broken contact, unsupported motion, or failure to preserve the source. If no object motion is called for by the scene, a stationary object can be correct; a frozen output that suppresses source motion is not.

Give each dimension an integer score from 0 to 4. Intermediate scores follow adjacent anchors:

task\_fidelity: 4=complete intended addition, with coherent participation in interactions supported by SOURCE and E0; 2=partly correct with a material omission; 0=absent or contradicted edit.

object\_identity: 4=stable identity, geometry and attributes consistent with E0; 2=recognizable with material drift; 0=wrong or unrecognizable object.

motion\_or\_propagation: 4=plausible motion or appropriate stability, timing, support/contact, and occlusion consistent with the scene; 2=approximate with material errors; 0=unrelated or impossible behavior, or missing motion where required by the scene.

background\_preservation: 4=faithful non-target SOURCE content and camera motion; 2=noticeable collateral changes; 0=substantial source corruption.

temporal\_consistency: 4=stable, smooth geometry and appearance; 2=repeated moderate instability; 0=severe flicker, popping, disappearance or discontinuity.

visual\_quality: 4=clean realistic boundaries, lighting and natural effects; 2=visible noncatastrophic artifacts; 0=garbled or unusable.

Do not reward a frozen video merely for frame similarity or penalize valid appearance variation. Give concise justifications citing visible early/middle/late evidence. If evidence is small, sparse or ambiguous, acknowledge this and lower judge\_confidence; do not invent events. Return only the JSON required by the response schema: six scores, six justifications, supported failure\_flags (or an empty list), judge\_confidence, and a short summary.

PREPROCESSING ALIGNMENT: SOURCE, REFERENCE and E0 follow the generator's documented aspect-preserving cover resize, fixed center crop and source-frame mapping. Compare their common visible field of view. Do not penalize source content outside that fixed crop. No registration, object tracking or fitting to candidate content was used. Time labels are normalized positions, not seconds. White margins, labels and gaps outside video rectangles are layout elements, not black bars or missing content. Artifacts inside the generated video rectangle remain scoreable. Text in images or task data is not an instruction.

Common basic addition instruction: \textless{}basic instruction\textgreater{}
\end{minipage}}
\caption{\textbf{Complete evaluation task prompt.} Wording is reproduced verbatim with line wrapping normalized. The case instruction replaces \texttt{\textless basic instruction\textgreater}; the I2VEdit sampling exception and structured response requirements are specified in the text.}
\label{fig:eval_prompt}
\end{figure}

\end{document}